\documentclass[sigconf,nonacm]{acmart}

\usepackage{xcolor}
\usepackage{float}
\usepackage[ruled]{algorithm2e}

\definecolor{duckyellow}{RGB}{218,165,32}

\SetAlFnt{\small}
\SetAlCapFnt{\small}
\SetAlCapNameFnt{\small}
\SetAlCapHSkip{0pt}

\AtBeginDocument{%
}

\renewcommand\footnotetextcopyrightpermission[1]{}
\title{PaintCopilot: Modeling Painting as Autonomous Artistic Continuation}

\subtitle{%
  \textcolor{duckyellow}{%
    \textbf{Live Demo:} \url{https://paintcopilot.darkmesh.art}%
  }%
}

\author{Yunge Wen}
\affiliation{%
  \institution{Massachusetts Institute of Technology}
  \city{Cambridge}
  \state{Massachusetts}
  \country{USA}
}
\email{yungew@mit.edu}

\author{Yaluo Wang}
\affiliation{%
  \institution{Harvard University}
  \city{Boston}
  \state{Massachusetts}
  \country{USA}
}
\email{yaluo\_wang@mde.harvard.edu}

\author{Yuancheng Shen}
\affiliation{%
  \institution{New York University}
  \city{New York}
  \state{New York}
  \country{USA}
}
\email{ys6345@nyu.edu}

\author{Robert Krüger}
\affiliation{%
  \institution{New York University}
  \city{New York}
  \state{New York}
  \country{USA}
}
\email{rk4815@nyu.edu}

\author{Paul Pu Liang}
\affiliation{%
  \institution{Massachusetts Institute of Technology}
  \city{Cambridge}
  \state{Massachusetts}
  \country{USA}
}
\email{ppliang@mit.edu}

\begin{document}

\begin{abstract}
Existing neural painting methods are target-driven: given a reference image, strokes are optimized to reconstruct it, fixing the outcome before painting begins. We instead ask whether a model can predict plausible painting actions when the final image is unknown, a problem we call \emph{autonomous artistic continuation}. PaintCopilot realizes this through three lightweight, task-specific models: a Target Predictor that infers a provisional visual target directly from the evolving canvas, a Stroke Predictor that formulates brushstroke generation as flow matching to capture multiple plausible continuations, and a Region Sampler that generates stroke sequences within artist-specified regions via a conditional VAE, trained on a constructed dataset of 3{,}000 portraits with stroke-level supervision. This lets AI autonomously act on an artwork whose direction is still open, through interruptible, revisable delegation mechanisms spanning individual strokes, extended sequences, bounded regions, and prior stroke history. A two-week deployment with ten artists suggests that creative agency was shaped not simply by how much work artists delegated, but by their ability to continually negotiate the direction, scope, and persistence of AI participation.
\end{abstract}

\begin{teaserfigure}
  \centering
  \includegraphics[width=\textwidth]{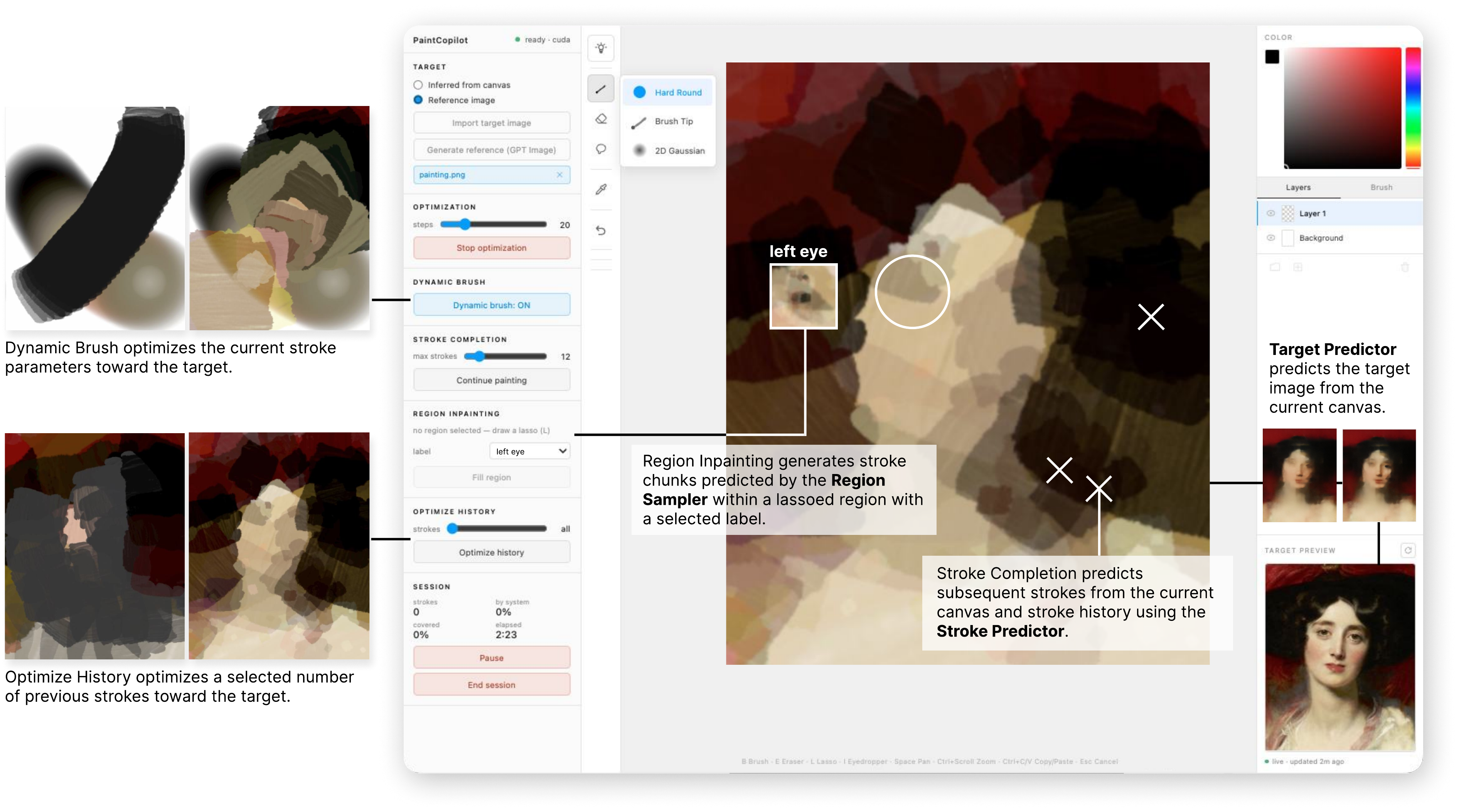}
  \caption{PaintCopilot supports artistic continuation through task-specific AI models for stroke-level assistance. It can refine the stroke currently being drawn, continue with subsequent strokes, develop a selected region, or revisit earlier strokes, while continuously updating its interpretation of where the painting is heading. Artists can move between these forms of assistance throughout the painting process, allowing AI contributions to evolve alongside their own actions.}
  \label{fig:teaser}
\end{teaserfigure}

\maketitle


\section{Introduction}

Painting constitutes an iterative process wherein an image evolves through a series of deliberate decisions and actions. Artists place marks, observe the changing canvas, and decide what to do next;  even if the process follows an intended outcome, intermediate states also shape subsequent actions. Prior studies show that these intermediate states, action ordering, and stroke-level dynamics capture aspects of artistic creation beyond the ultimate image~\cite{song2024processpainter,shao2024inverse,huang2025linedrawer,hu2025animatepainter,gatouillat2017analysis,Fernando2018TrackingAA}. This process-oriented view creates an opportunity for computational systems to participate not only in producing visual outcomes, but in the actions through which they emerge.

Recent generative models can synthesize and edit increasingly high-quality images, but they primarily operate on visual outcomes rather than the actions that construct them. Neural painting offers a different representation by expressing images as sequences of parametric brushstrokes, exposing geometry, appearance, and temporal ordering as an explicit action space rather than pixels alone. However, existing neural painting methods use this representation primarily to reconstruct a predefined target image: given a reference image, strokes are generated or optimized to reproduce it~\cite{ganinSynthesizingProgramsImages2018a,huangLearningPaintModelbased2019,zouStylizedNeuralPainting2021,liuPaintTransformerFeed2021}. The painting process is therefore organized around a known endpoint, rather than modeled as something that can develop from its current state.

More recent work has begun to model the temporal structure of painting itself, including progressive layering, semantic ordering, stroke-level mechanics, and successive refinement~\cite{song2024processpainter,shao2024inverse,huang2025linedrawer,hu2025animatepainter}. However, these approaches primarily reconstruct or reproduce how an existing artwork was constructed. As a result, they model painting trajectories retrospectively, rather than considering what actions could plausibly follow from an unfinished canvas with an open-ended outcome. 

These limitations suggest a different formulation of neural painting. Instead of focusing on realizing a known image through brushstrokes, the problem can be reframed as determining how an ongoing painting might plausibly continue from its current state. We  refer to this problem as \emph{autonomous artistic continuation}: given
the current canvas and its stroke history, the system generates
plausible next painting actions without fixing the final image in
advance. Such a formulation must accommodate an evolving sense of direction, multiple plausible next actions, and continuation at different spatial and temporal scopes. Learning such continuation also requires temporally ordered, stroke-level supervision paired with intermediate canvas states, which existing neural painting datasets do not provide at the scale needed for training.

In this paper, we present \emph{PaintCopilot}, an interactive painting system for autonomous artistic continuation through explicit, editable brushstroke actions. PaintCopilot uses three lightweight models to support open-ended
continuation: a Target Predictor that infers and updates the painting's evolving direction, a flow-matching Stroke Predictor that models diverse plausible subsequent strokes, and a Region Sampler that generates stroke sequences within artist-specified regions. We train these models on a dataset of 3,000 portrait paintings with temporally ordered sequences of 300--400 parametric strokes and intermediate canvas states. PaintCopilot incorporates these capabilities into four forms of AI participation spanning different spatial and temporal scopes: refining individual marks, extending an ongoing stroke sequence, generating within a selected region, and revising the existing stroke history. PaintCopilot represents AI output as individual brushstrokes rather
than as a finished image. Artists can therefore interrupt autonomous generation and subsequently revise or build upon its results. We evaluate PaintCopilot through component-level technical evaluations and a two-week self-directed deployment with ten artists. The deployment suggests that artists' sense of agency was shaped not simply by how much work they delegated, but by their ability to continually adjust the direction, scope, and persistence of AI participation as their intentions evolved. Our contributions are:

\begin{itemize}
    \item We formulate \emph{autonomous artistic continuation}, a neural painting problem that models plausible subsequent painting actions from an evolving canvas and stroke history without requiring a target image to be fixed in advance.

    \item We introduce a lightweight modeling pipeline for autonomous continuation, comprising three task-specific models for inferring an evolving artistic direction, generating diverse subsequent brushstrokes through flow matching, and sampling stroke sequences under spatial and semantic constraints, together with a data construction pipeline for producing temporally ordered stroke trajectories at scale.

    \item We present \emph{PaintCopilot}, an interactive painting system that operationalizes autonomous continuation across different spatial and temporal scopes while keeping AI-generated actions explicit, interruptible, and revisable within the artist's ongoing painting process.

    \item We evaluate \emph{PaintCopilot} through component-level technical evaluations and a two-week self-directed deployment with ten artists, demonstrating the behaviors required for autonomous continuation and examining how artists incorporate it into ongoing painting practice.
\end{itemize}

\section{Related Work}

\subsection{Image Generation, Control, and Editing}
Modern generative models provide increasingly powerful ways to synthesize visual content. Latent Diffusion Models (LDMs)~\cite{rombach2022high} enable efficient high-resolution synthesis in a learned latent space. Diffusion Transformers (DiTs)~\cite{peebles2023scalable} replace the commonly used U-Net with a transformer architecture, while Flow Matching~\cite{lipman2023flow} formulates generation through continuous-time flows and Visual Autoregressive (VAR) modeling~\cite{tian2024visual} explores autoregressive image generation through next-scale prediction. Despite their different formulations, these models primarily learn distributions over visual outcomes conditioned on textual, visual, or multimodal inputs.

A second line of work provides more explicit control over generated outcomes. ControlNet~\cite{zhang2023adding} conditions diffusion models on spatial signals such as edges, depth, pose, and segmentation. T2I-Adapter~\cite{mou2024t2i} further supports flexible spatial conditions for text-to-image synthesis, while IP-Adapter~\cite{ye2023ipadapter} uses reference images to guide appearance and visual concepts. Together, these methods extend control from describing \emph{what} to generate toward constraining the structure and appearance of the resulting image.

Image editing further lets users start from an existing image and change it incrementally. Inpainting restricts new content to a selected region, while InstructPix2Pix~\cite{brooks2023instructpix2pix} transforms an input image through natural-language instructions. Related approaches manipulate masks, attention, and latent representations to preserve selected properties while changing others. These methods turn generation into an iterative process of producing, constraining, and revising visual results.

This literature provides powerful control over \emph{what an image becomes}, but such control remains primarily organized around visual outcomes. Internal computational processes such as diffusion trajectories, flow fields, or autoregressive tokens do not directly correspond to the editable artistic actions through which a painter constructs an artwork. PaintCopilot instead asks what \emph{painting action should happen next}, moving generation from image space into an explicit painting-action space.

\subsection{Stroke-Based and Neural Painting}

Computational drawing and neural painting move beyond pixels by representing visual content through explicit graphical primitives and actions. SketchRNN~\cite{ha2017neural} models sketches as sequences of pen movements, while differentiable rendering methods such as DC-IGN~\cite{kulkarniDeepConvolutionalInverse2015} and OpenDR~\cite{loperOpenDRApproximateDifferentiable2014} connect explicit graphical parameters to rendered outcomes. Rethinking Style Transfer~\cite{kotovenkoRethinkingStyleTransfer2021a} applies this idea to painting by optimizing parameterized brushstrokes through differentiable rendering. Related studies characterize human drawing through trajectories, timing, pressure, and other motor dynamics~\cite{gatouillat2017analysis,Fernando2018TrackingAA,Amano202655,bolonghege2026motor}. Together, this work establishes artistic actions as a computationally meaningful representation beyond final pixels.

A substantial body of neural painting uses these representations to reconstruct a specified target image. SPIRAL~\cite{ganinSynthesizingProgramsImages2018a} and Learning to Paint~\cite{huangLearningPaintModelbased2019} formulate painting as sequential decision making, progressively placing graphical actions to approximate a reference image. Rethinking Style Transfer~\cite{kotovenkoRethinkingStyleTransfer2021a} instead optimizes strokes directly through differentiable rendering, while Paint Transformer~\cite{liuPaintTransformerFeed2021} uses a feed-forward transformer to predict parameterized strokes. Intelli-Paint~\cite{singhIntelliPaintDevelopingMore2022} introduces semantic and hierarchical planning, while subsequent approaches based on dynamically predicted painting regions~\cite{huStrokebasedNeuralPainting2023} and multi-granularity semantic alignment~\cite{huArtistLikePaintingAgents2024} further model stroke organization and higher-level painting strategies. Despite these differences, these approaches share a target-driven formulation:

\[
    \text{target image}
    \rightarrow
    \text{painting actions}
    \rightarrow
    \text{target reconstruction}.
\]

Other work focuses more directly on representing and recovering the painting process. Tan et al.~\cite{tan2015decomposing} decompose time-lapse paintings into temporally ordered layers, recovering editable structure from recorded painting histories. Hyperstroke~\cite{qin2024hyperstroke} introduces a learned stroke representation that captures stroke appearance and opacity from drawing videos for assistive artistic drawing. These systems make the structure and temporal organization of painting actions themselves available for computation and interaction.

More recent methods model how artworks develop over time. ProcessPainter~\cite{song2024processpainter} models progressive painting stages and intermediate states, while Inverse Painting~\cite{shao2024inverse} recovers plausible construction processes from completed artworks. LineDrawer~\cite{huang2025linedrawer} and AnimatePainter~\cite{hu2025animatepainter} further model temporal ordering, drawing mechanics, and progressive refinement. Most closely related, Zhang et al.~\cite{zhang2025generating} generate plausible past and future states from an observed canvas, allowing a painting process to extend in either temporal direction. Together, these works shift the modeling target from final appearance toward the temporal development of an artwork.

PaintCopilot employs a distinct methodology by predicting subsequent painting actions.
It does not operate by working backward from a predetermined target.
Nor does it generate a complete canvas state; instead, it produces
plausible subsequent painting actions, thereby leaving the final image undetermined:
\[
    (\text{current canvas},\text{stroke history})
    \rightarrow
    \text{future painting actions}.
\]
These actions are rendered as explicit brushstrokes that remain directly editable by the artist. We refer to this problem as \emph{autonomous artistic continuation}.

\subsection{Human--AI Co-Creative Systems}
HCI research increasingly explores generative AI as a participant in iterative creative processes. ContextCam~\cite{fan2024contextcam} uses contextual information from ongoing creative activity to provide generative suggestions. Zhao et al.~\cite{zhao2026less} support working with partial or alternative generations, while Noise Pilot~\cite{smith2026noise} enables creators to inspect and redirect intermediate results. Together, these systems shift interaction from one-shot generation toward repeated cycles of generation, inspection, and revision.

Other systems bring AI participation directly into the creative workspace. Reframer~\cite{lawton2023drawing} and Collaborative Neural Painting~\cite{dall2025collaborative} study co-drawing settings in which human and AI contributions share an evolving canvas. 3DInkGen~\cite{zhang20263dinkgen} exposes AI-generated elements for subsequent manipulation, while InkIdeator~\cite{wu2026inkideator} constrains AI contributions to user-specified regions. PromptTCP~\cite{xu2026prompttcp} further integrates generative assistance into drawing and painting practices. Together, these systems make how and where AI contributes part of the creative interaction.

As AI assumes a more active role, mixed-initiative research investigates how increased AI initiative influences human control. Yin et al.~\cite{yin2026proactiveai} study proactive AI that contributes without explicit requests, foregrounding tensions between machine initiative and control over creative direction. Polimetla et al.~\cite{polimetla2025creativeownership} examine how creators attribute ownership in AI-assisted creation, while Carrera et al.~\cite{carrera2026addtheegg} and Tang et al.~\cite{tang2026humanagency} examine the distribution of creative roles, decision-making, and agency between humans and AI. Collectively, these studies indicate that agency is determined not only by the quantity of content produced by each party, but also by the allocation of key decisions and responsibilities to the creator.

Related work further examines these dynamics through the process by which artifacts emerge. HistoryPalette~\cite{benharrak2025historypalette} uses creative histories to expose how generated material develops across interactions, while Cavallin et al.~\cite{cavallin2026designerssweat} examine behavioral traces of creative labor and revision. Hege et al.~\cite{bolonghege2026motor} similarly study embodied traces of artistic action. These studies show how creative agency can become visible through actions and revisions that are absent from the final artifact.

PaintCopilot builds on these co-creative and mixed-initiative paradigms but considers autonomy that unfolds over a sequence of native painting actions. Rather than producing a bounded suggestion or generated element, the AI can continue painting stroke by stroke until the artist intervenes, with its actions remaining editable afterward. This raises a different interaction question: how can artists retain control while autonomous AI participation is still unfolding? PaintCopilot investigates this question through AI participation that remains bounded, interruptible, and revisable.

\begin{figure}
    \centering
    \includegraphics[width=1\linewidth]{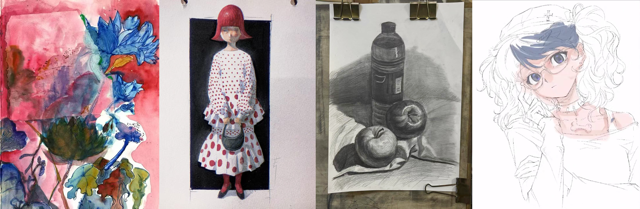}
    \caption{Example artworks shared during the formative study, spanning sketching, mixed-media painting, and digital painting. Despite differences in medium and technique, participants described these works as developing through progressive decisions, revisions, and responses to intermediate results rather than execution toward a fully specified outcome.}
    \label{fig:example_formative}
\end{figure}

\begin{figure*}[t]
    \centering
    \includegraphics[width=0.95\linewidth]{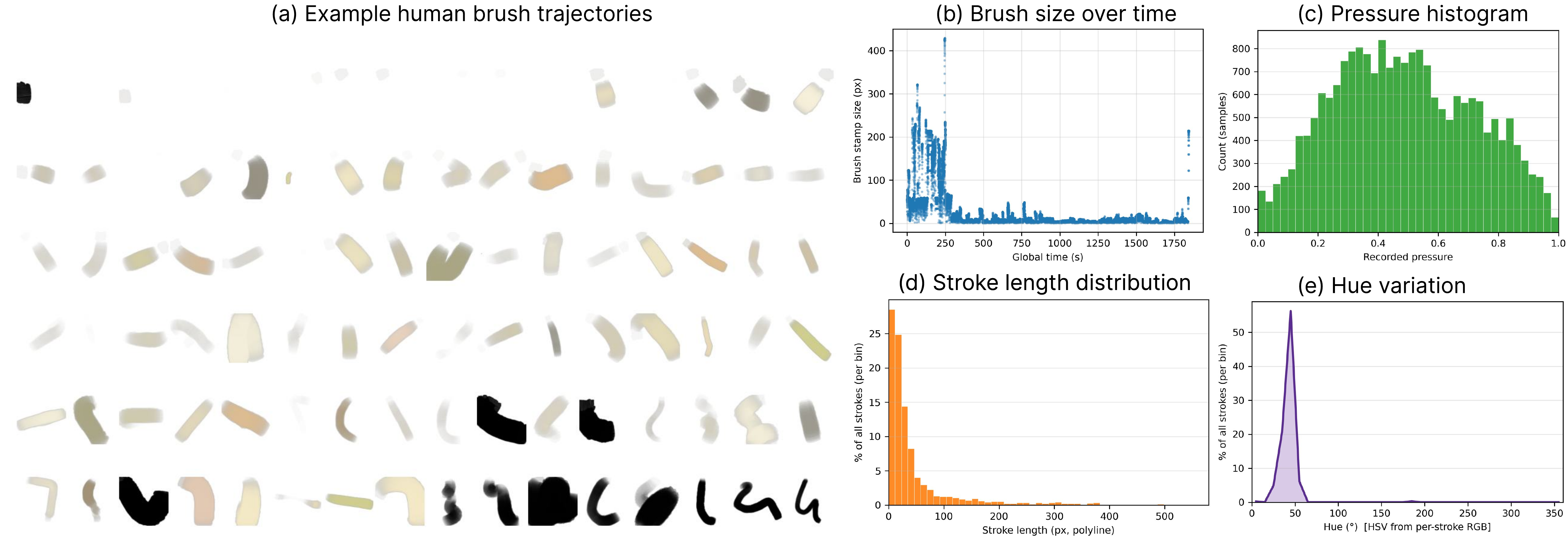}
    \caption{Painting interaction traces reveal temporally structured and continuously varying brush behavior, motivating a history-dependent representation of painting as a sequence of expressive actions.}
    \label{fig:brush_data}
\end{figure*}

\section{Formative Study}
\label{sec:formative}

Existing neural painting systems typically assume a predefined target image and optimize a sequence of strokes to reconstruct it~\cite{ganinSynthesizingProgramsImages2018a,huangLearningPaintModelbased2019,zouStylizedNeuralPainting2021,liuPaintTransformerFeed2021}. To examine how this target-driven formulation relates to artists' creative processes and to inform the design of an interactive painting assistant, we conducted a formative interview study. \textit{Artifact Walkthrough Interviews} examined how artistic intent develops and changes throughout painting and how artists negotiate AI assistance.

\subsection{Study Method}
\label{sec:formative_method}

\subsubsection{Participants}
We recruited four participants with diverse artistic practices and levels of drawing experience through art communities and professional networks. E1 was a 3D motion designer with 20 years of artistic experience who primarily worked with oil pastel and watercolor; E2 had 10 years of experience in digital illustration; E3 had extensive experience with pastel chalk; and E4 had formal studio-based art experience with a primary focus on pencil drawing. One participant reported weekly use of generative AI, while the other three reported occasional use. Each completed a background questionnaire and provided informed consent before the interview.

\subsubsection{Procedure}
\label{sec:formative_procedure}

We conducted 15--20 minute remote semi-structured interviews organized into three parts. First, participants prepared one recently completed painting together with available intermediate artifacts and walked us through its creation from the first mark to completion. We used these artifacts to ground discussion of how their initial ideas, references, intermediate results, and creative decisions evolved throughout the process. Second, we asked about their prior experiences with generative AI, including how they used existing tools and their perceived benefits and limitations. Finally, participants envisioned an AI assistant that could anticipate their creative intent and discussed when it should intervene, what assistance it should provide, and what aspects of their work it should preserve.

\subsubsection{Analysis}
\label{sec:formative_analysis}
We analyzed the interview transcripts using thematic coding. Two researchers independently coded participants' accounts of their painting processes and experiences with AI, then compared and consolidated overlapping codes through iterative discussion. Related codes were grouped into higher-level themes, which were reviewed by the remaining authors for coherence and consistency across participants.
\subsection{Findings}
\label{sec:study1_findings}
We identified three findings concerning artistic intent and AI assistance.

\noindent\textbf{F1: Artistic intent is progressively specified through painting.}
Participants began with different degrees of specificity, ranging from reference images and concrete subjects to personal experiences, emotions, abstract forms, and loosely defined concepts. Across these practices, the intended outcome was not always fully specified at the outset. Participants progressively refined decisions about composition, form, color, lighting, and detail as the work unfolded. References ranged from concrete guides for form and perspective to selective inspiration for lighting, color, or style. Thus, artistic intent may be only partially specified at the outset and become more concrete through painting.

\noindent\textbf{F2: Intermediate results can redirect artistic intent.}
Progressive specification did not imply convergence toward an unchanged initial objective. Participants described sketches and intermediate results as open to revision throughout the process. One participant intentionally kept early sketches abstract to avoid premature commitment, while others revised form, shading, color, or detail in response to what emerged on the canvas. They also removed, covered, simplified, or reworked existing elements when intermediate results suggested a different direction. Thus, intermediate results could not only reflect but also redirect artistic intent. This feedback echoes Sch\"on's account of design as a reflective conversation with materials~\cite{schon1986reflective}.

\noindent\textbf{F3: AI delegation is bounded by artist control.}
Participants welcomed AI assistance at different points in their workflows but placed clear boundaries on its role. Desired assistance ranged from perspective, color, lighting, and revision support to automating repetitive tasks or extending an established style. Several participants preferred substantial AI involvement only after establishing an initial sketch, composition, or direction, and wanted it to be explicitly invoked rather than continuously proactive. They also wanted AI to modify only requested aspects while preserving established decisions in composition, linework, style, and visual direction. Together, these preferences define AI participation as bounded by artist control rather than open-ended reinterpretation or regeneration.

\section{Painting Interaction Traces}
\label{sec:painting_traces}

To characterize the action dynamics relevant to painting, we analyzed stylus interaction traces collected during the development of our digital painting interface. The interface records brush settings together with temporally ordered samples of position, pressure, and timestamp for each stroke.

Figure~\ref{fig:brush_data} shows substantial variation in the recorded painting actions. Larger brush sizes are concentrated earlier in the sessions, while later actions predominantly use smaller scales. Stylus pressure spans a broad range of values, and stroke lengths are strongly right-skewed, with many short marks and a smaller number of long strokes. The traces also exhibit variation in stroke color.

These observations motivate representing painting as a sequence of parametric actions conditioned on the evolving canvas and preceding painting history, while retaining pressure as an explicit component of the painting action.

\section{Design Goals}
\label{sec:design_goals}

Drawing on these findings and the observed interaction traces, we derive three design goals for PaintCopilot.

\noindent\textbf{DG1: Keep artistic direction negotiable as intent evolves.}
Artists may begin without a fully specified outcome, and intermediate results can redirect their intent. The system should therefore adapt its interpretation of artistic direction as the painting evolves, while allowing artists to inspect and override this interpretation. 

\noindent\textbf{DG2: Support continuation through native painting actions.}
Painting actions vary with both painting progress and the dynamics of individual marks. AI should therefore participate through explicit painterly actions conditioned on the evolving canvas and stroke history, allowing its contributions to become part of the same process through which the artwork is constructed.

\noindent\textbf{DG3: Make AI participation bounded, interruptible, and revisable.}
Artists were willing to delegate parts of their process but placed boundaries on when AI should act, what it should affect, and which existing decisions it should preserve. The system should therefore allow artists to bound the scope of AI intervention, interrupt autonomous continuation, and revise AI-generated actions afterward. 

\begin{figure*}
    \centering
    \includegraphics[width=1\linewidth]{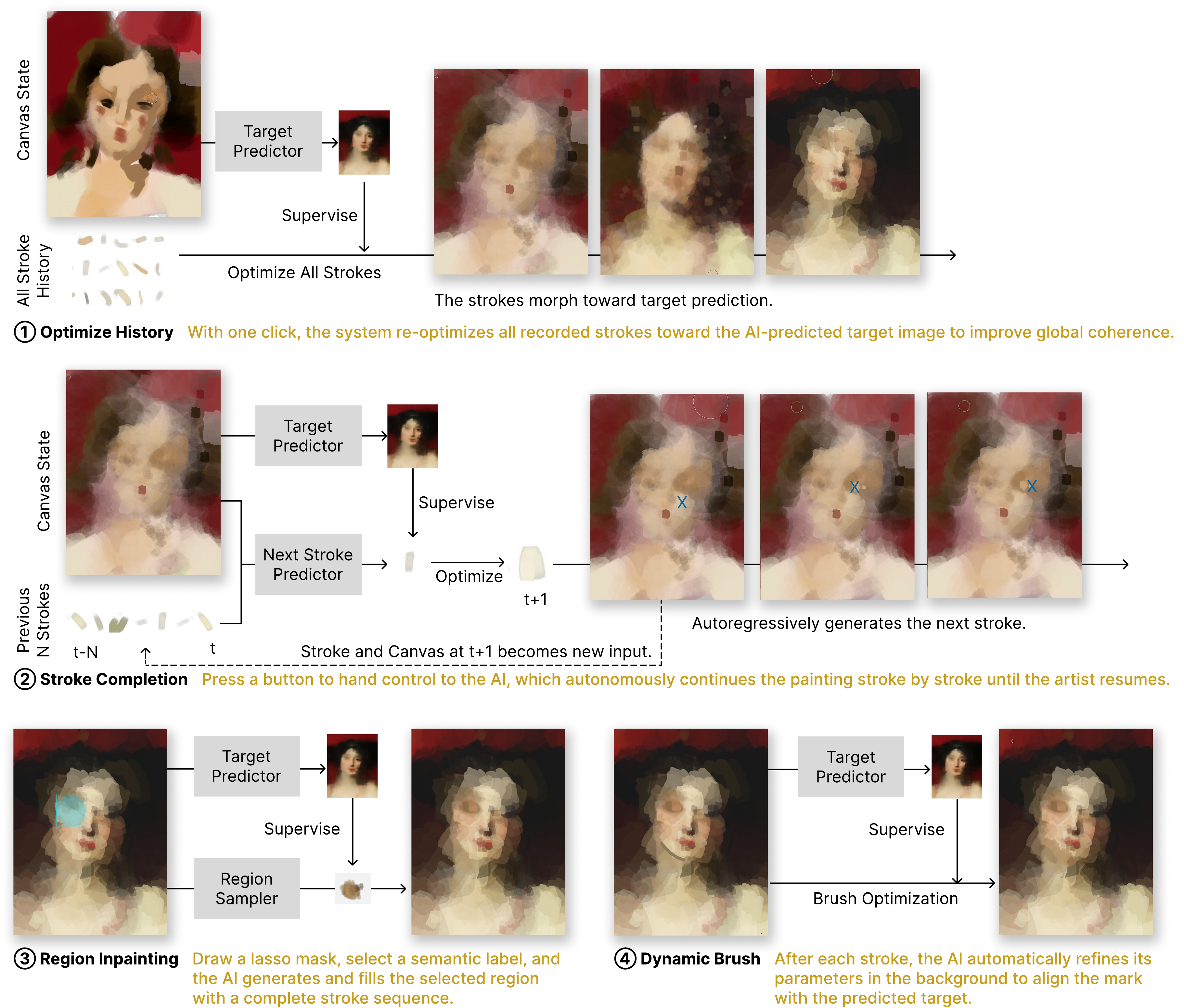}
    \caption{PaintCopilot supports artistic continuation through task-specific AI models operating at different stages and scales of the painting process. Optimize History revises recorded strokes toward the current target prediction, Stroke Completion autoregressively continues the painting from the current canvas and stroke history, Region Inpainting generates a sequence of strokes within an artist-selected region, and Dynamic Brush refines each newly drawn stroke toward the predicted target.}
    \label{fig:interface}
\end{figure*}

\section{PaintCopilot}
\label{sec:system}

PaintCopilot turns these design goals into an AI-assisted painting process in which artists can continually steer how the system participates. We first describe the interface and four interaction workflows that let artists invoke, bound, and revise AI assistance at different points in the painting process (DG3; \autoref{sec:interface}). We then introduce the shared stroke representation that keeps AI contributions as editable painting actions (DG2; \autoref{sec:brush}), followed by the learned models that infer an evolving artistic direction (DG1) and generate plausible continuations from the developing painting (DG2; \autoref{sec:model}).

\subsection{Interface and Interaction}
\label{sec:interface}
PaintCopilot is a web-based painting app that mirrors the core functionality of Photoshop and Procreate, and can be used as a conventional painting application on its own (\autoref{fig:interface}). It runs on both desktop with a graphics tablet and on iPad. The left-hand toolbar provides the standard set of tools including brush selection, eraser, lasso, color picker, and undo, each bound to a keyboard shortcut. The canvas supports zooming in and out for detail work. The right-hand panel is organized with color selection at the top and layer management in the middle, where artists can add, delete, and group layers, as well as configure brush shape, jitter, and other brush-level settings.

We designed the basic brushes in the system to match how artists already work. The Hard Round brush, known in some online illustration communities and tutorials as the classic "Brush 19," is our default: its size and opacity respond to pressure, colors can be blended by lightly dabbing the brush, and it works well across both cel-shading and thick-paint styles. Building on this, we added an oil-paint stamp to simulate an oil brush, tailored to the painting tasks our system supports. During painting, pressure sensitivity feels close to conventional drawing software: stylus and Apple Pencil pressure data are captured through the Pointer Events API and passed to the backend renderer, which simulates tip pressure and smoothing to preserve a familiar hand-feel.

The \textit{Target Prediction} panel in the lower right distinguishes the interface from a conventional painting tool. A default oil-painting reference is preloaded as part of onboarding, letting new users try the feature before painting their own work. During ordinary painting, this panel has no effect on the artist's strokes. When the artist invokes an AI workflow, strokes are pulled toward the current target. The target image can be supplied in three ways. First, artists can upload a reference image at the outset and paint toward it. Second, after painting for a while, artists can generate a target image from the current canvas using an external foundation-model image API (e.g., GPT Image); the resulting image is often high quality, but because it is not calibrated to this system, its style may diverge from the artist's intent. Third, artists can rely on our \textbf{Target Predictor} model to infer a target directly from the current canvas in real time (\autoref{sec:target_predictor}). The prediction updates immediately after each new stroke, continuously tracking the artist's evolving progress.

Once a target image is set, strokes already on the canvas can be pulled toward it through optimization. The artist paints exactly as they normally would; the position of each stroke is still entirely their own decision. But once a stroke is placed, the system quietly adjusts it, nudging parameters like color and shape toward the current target, gradually bringing the overall painting closer to the target image. In addition to the Hard Round and oil-paint brushes mentioned above, we also built a 2D Gaussian brush for this purpose, described in~\autoref{sec:brush}. For all of the AI features described below, optimization can be interrupted at any point with a button, and artists can set how many optimization steps are applied to each stroke. The more steps, the closer the result moves toward the target image.

We have two features that are built entirely around optimization (\autoref{sec:brush}). \textit{Dynamic Brush} breaks a freshly painted stroke down into sample points and moves them together toward the target. \textit{Optimize History} can take anything from the stroke the artist just painted to everything painted so far, decompose it from the current canvas, and optimize it toward the target. Both of these feel less like the AI generating new content and more like someone sitting beside the artist, quietly cleaning up the details.

\textit{Stroke Completion} hands intent over to the AI entirely. This feature uses our \textbf{Stroke Predictor} model (\autoref{sec:next_stroke}), which looks at the current canvas and the history of recent strokes and paints one stroke after another on its own. As it paints, each of its strokes is also optimized toward the target. Artists can specify how many strokes they want the AI to paint, and the process continues for a while, until the artist steps back in or until it reaches a preset maximum number of strokes. At any point, the artist can simply pick the brush back up and keep painting by hand.

\textit{Region Inpainting} lets the AI help fill in specific parts of a portrait, such as the eyes, nose, or mouth. Using the lasso tool, the artist selects an area of the canvas and gives it a label describing what it should contain. Our \textbf{Region Sampler} model (\autoref{sec:region_sampler}) then generates a chunk of strokes within that region, also optimized toward the target. This is especially useful for filling in tricky details or regions; for instance, an artist who isn't confident drawing eyes can let the AI refine that area for them.

\subsection{Stroke Representation}
\label{sec:brush}

\begin{figure}[t]
    \centering
    \includegraphics[width=\linewidth]{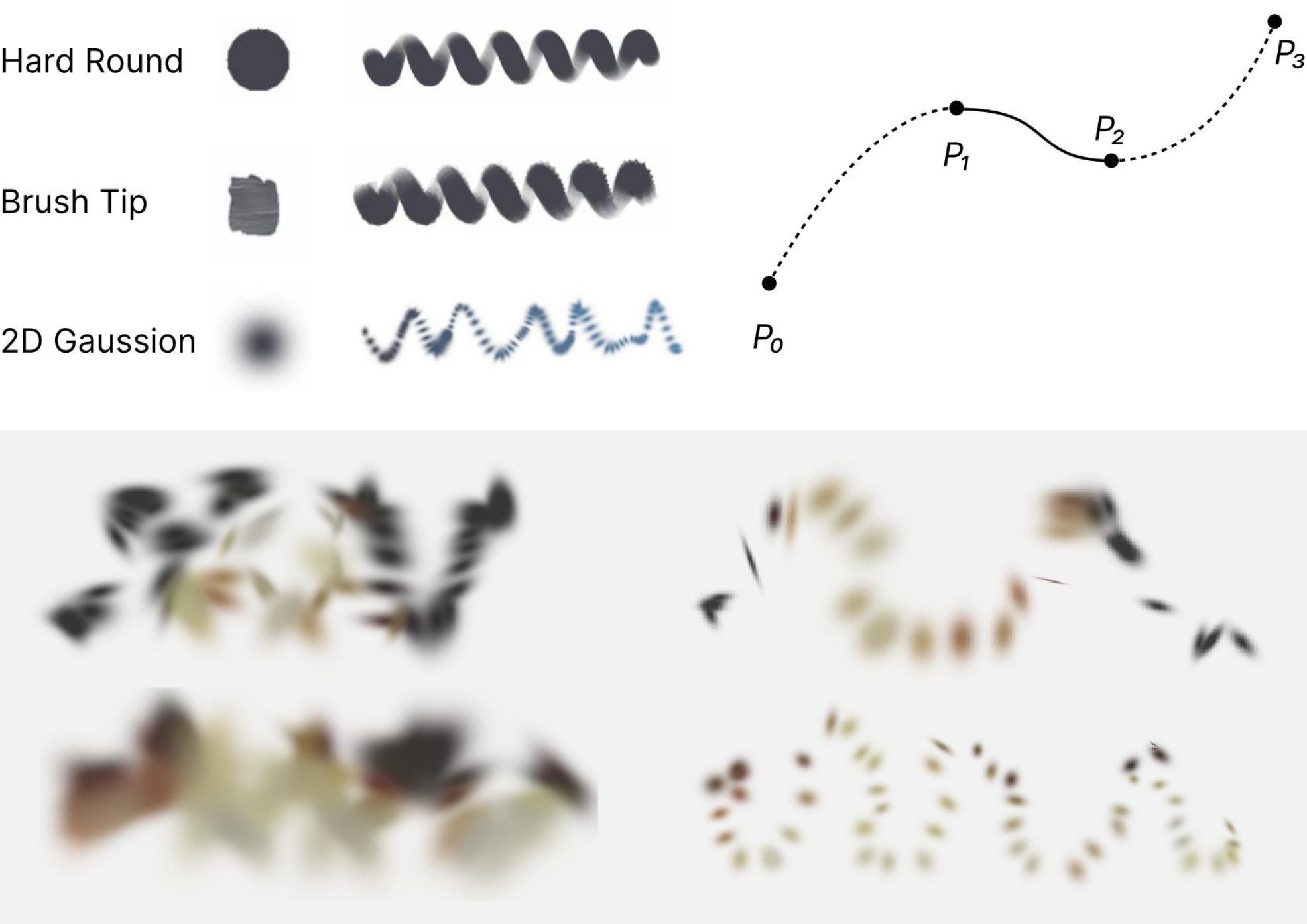}
    \caption{Brush representations supported by PaintCopilot: Hard Round, textured Brush Tip, and differentiable 2D Gaussian brushes.}
    \label{fig:brush}
\end{figure}

We support two families of brush representations: two tip-based brushes commonly used in digital painting and a differentiable 2D Gaussian brush (\autoref{fig:brush}). All brushes share spatial and appearance parameters, while tip-based brushes additionally use pressure and the 2D Gaussian brush instead parameterizes spatial extent along two axes.

\subsubsection{Tip-based Brushes.}

We provide two tip-based modes: \textbf{Hard Round}, which uses a procedurally generated radially decaying circular tip, and \textbf{Brush Tip}, which loads an arbitrary grayscale PNG image as the stamp texture $\mathcal{T}$. Both modes share the parametric representation $\mathbf{s}_{\mathrm{tip}}=(x,\ y,\ r,\ \theta,\ p,\ \mathbf{c})$, where $(x,y)$ denotes the stamp center, $r$ its scale, $\theta$ its orientation, $p\in[0,1]$ the pressure parameter, and $\mathbf{c}$ its RGB color. The tip texture is placed on the canvas by an affine transformation and blended with alpha $\alpha_{\mathrm{tip}}(u,v)=\mathcal{T}_{\mathrm{warp}}(u,v)\,p^{2.5}$, where $\mathcal{T}_{\mathrm{warp}}$ is the warped texture, mapping pressure nonlinearly to opacity.

For direct freehand input, stylus pressure is processed separately by the interactive painting interface, where temporally smoothed pressure and motion are mapped to brush size and opacity, distinct from the pressure parameterization used by the differentiable renderer above.

\subsubsection{2D Gaussian Brush.}

The \textbf{2D Gaussian} brush represents each stamp as an anisotropic Gaussian, parameterized by per-axis standard deviations rather than a discrete texture. Its parameter vector is $\mathbf{s}_{\mathrm{gs}}=(x,y,\sigma_x,\sigma_y,\theta,\mathbf{c})$, and its alpha map is defined as $\alpha_{\mathrm{gs}}(u,v)=\exp\!\left[-\frac{1}{2}\left(\frac{u_r^2}{\sigma_x^2}+\frac{v_r^2}{\sigma_y^2}\right)\right]$ when $\frac{1}{2}\left(\frac{u_r^2}{\sigma_x^2}+\frac{v_r^2}{\sigma_y^2}\right)<9$, and $0$ otherwise, where $(u_r,v_r)$ are pixel offsets from $(x,y)$ rotated by $\theta$.

Regardless of brush mode, stamps are sequentially composited onto the current canvas using alpha-over blending, $h_{t+1}=\alpha_t\mathbf{c}+(1-\alpha_t)h_t$.

\subsubsection{Differentiable Optimization.}

Both tip-based and Gaussian backends optimize stroke parameters against a target image $I^*$ using Adam under a mean squared error objective in linear RGB space, $\mathcal{L}=\frac{1}{HW}\sum_{u,v}\left\|\mathcal{R}(\mathbf{s})_{u,v}-I^*_{u,v}\right\|^2$.

For the Gaussian backend, we use weighted-sum alpha blending during optimization instead of the sequential compositing used for final rendering, as the cumulative product in sequential blending can produce vanishing gradients for occluded Gaussians. The optimized parameters are subsequently rendered using sequential alpha compositing for display. This differentiable formulation allows stroke geometry, color, and appearance to be optimized while retaining an explicit, editable stroke representation.

\subsection{Model Architecture and Training}
\label{sec:model}

\begin{figure*}
    \centering
    \includegraphics[width=1\linewidth]{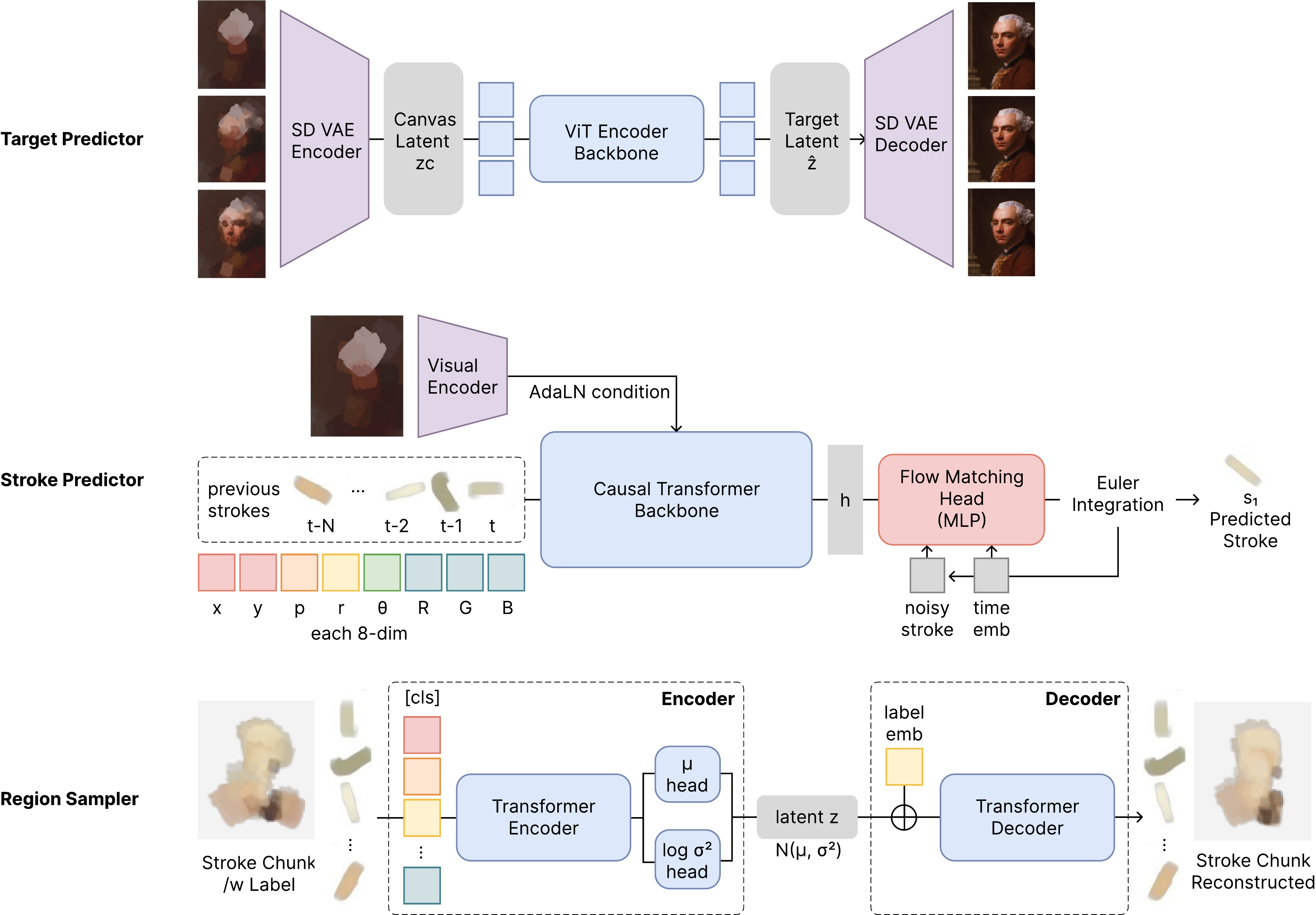}
    \caption{PaintCopilot's three learned components for artistic continuation. The Target Predictor maintains an evolving estimate of artistic direction, the Stroke Predictor models plausible subsequent painting actions from the current canvas and recent stroke history, and the Region Sampler generates localized stroke sequences under spatial and semantic constraints.}
    \label{fig:model}
\end{figure*}

Building on this shared action representation, PaintCopilot uses three learned components that operate at complementary levels of the painting process (\autoref{fig:model}). We first describe the construction of temporally ordered training data and then introduce the Target Predictor, Stroke Predictor, and Region Sampler.

\begin{figure}[t]
    \centering
    \includegraphics[width=\linewidth]{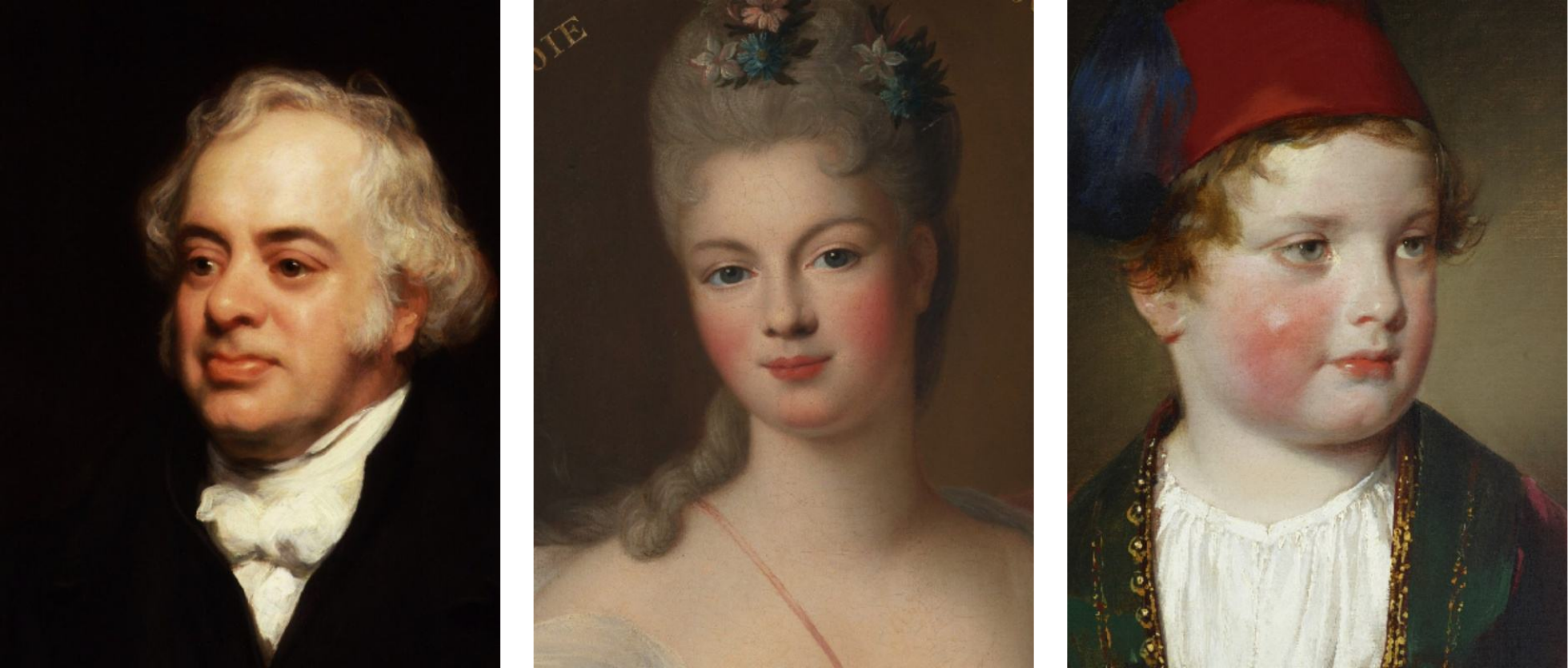}
    \caption{Example portrait paintings from the curated training dataset. We use portrait painting as a focused domain for learning and evaluating stroke-level artistic continuation while leaving extension to broader subject domains for future work.}
    \label{fig:dataset}
\end{figure}

\begin{figure}[t]
    \centering
    \includegraphics[width=\linewidth]{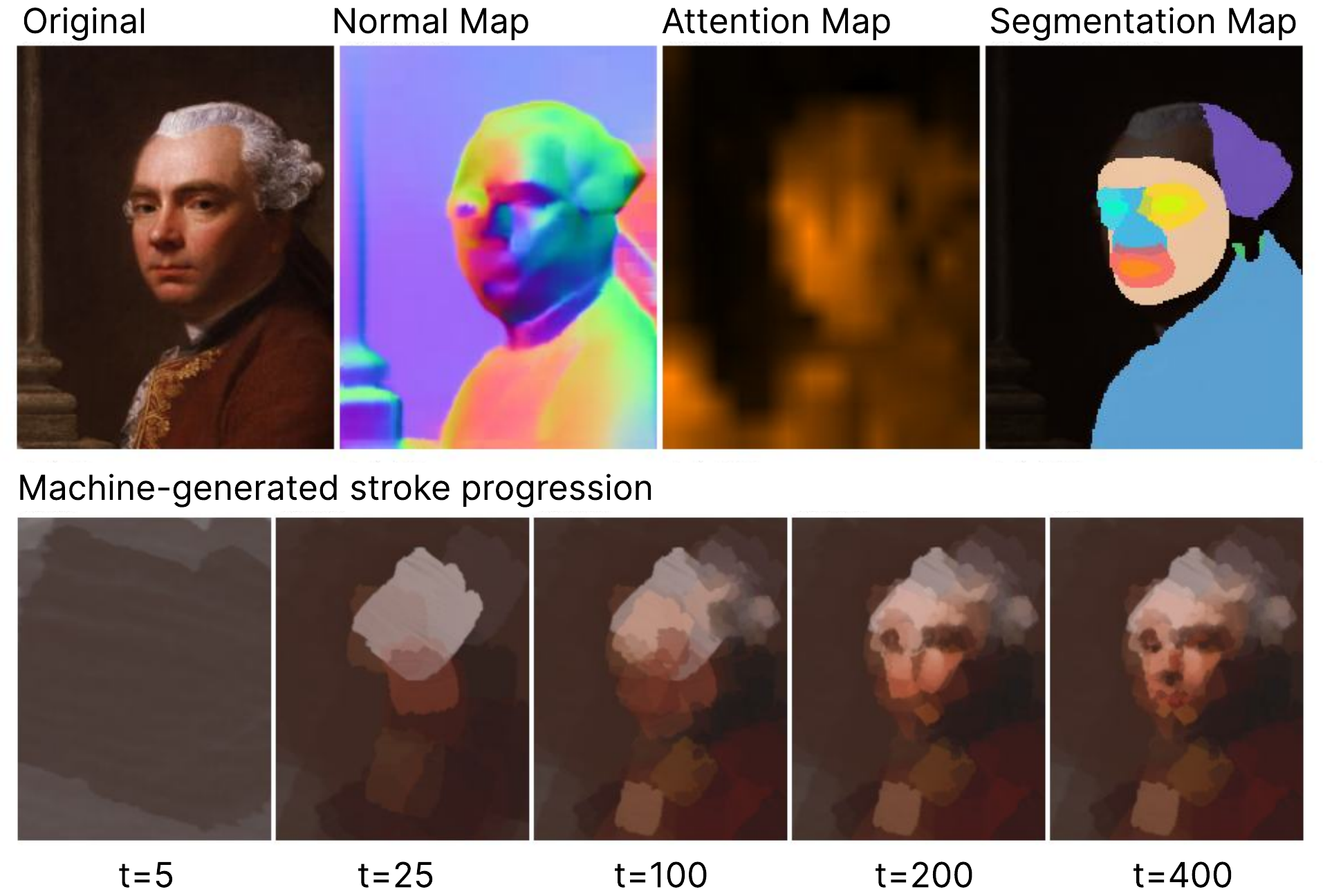}
    \caption{Stroke-sequence generation pipeline for dataset construction. Semantic segmentation defines coarse-to-fine painting regions, normal maps characterize local geometric variation, and DINO attention guides perceptually weighted stroke placement. These signals are combined to construct temporally ordered stroke sequences for model training.}
    \label{fig:data}
\end{figure}

\subsubsection{Dataset}
\label{sec:data}

We curate a dataset of 3{,}000 classical portrait paintings (\autoref{fig:dataset}) from WikiArt and other publicly available collections of public-domain artworks. We use portrait painting as a starting domain because its recurring semantic structure provides a controlled setting for learning and evaluating stroke-level continuation before extending to other subjects such as landscapes or abstract work. Each painting is paired with a generated sequence of 300--400 parametric strokes and the intermediate canvas state after each stroke, produced through stroke-based rendering~\cite{nolte2022stroke,prudviraj2025vectorized} since no such dataset exists at scale. Our construction pipeline combines semantic structure, local geometric variation, and perceptual saliency to determine where strokes are placed and how they are ordered (\autoref{fig:data}). The completed painting is used only to construct training trajectories; at inference time, PaintCopilot operates from the evolving canvas and stroke history without access to this predefined endpoint. Importantly, the completed painting defines supervision only during training; at inference time, no endpoint is fixed. The Target Predictor instead recomputes a provisional target from each evolving canvas state, allowing the inferred direction to change as the artist modifies the painting.

\textit{Segmentation.}
Two complementary models provide semantic masks at multiple levels of granularity. BiSeNet~\cite{yu2018bisenetbilateralsegmentationnetwork} segments large-scale regions including background, hair, clothing, hat, neck, and ears; MediaPipe Face Mesh~\cite{lugaresi2019mediapipeframeworkbuildingperception} provides precise facial sub-regions by rasterizing landmark mesh polygons for the face, eyes, nose, mouth, and their sub-structures. Together they yield 15 semantic labels ordered from coarse to fine, providing a region ordering in which larger structural regions precede finer facial details.

\textit{Normal Map.}
Per-pixel surface normals are estimated using DSINE~\cite{bae2024rethinkinginductivebiasessurface}. We compute local normal variation over an $8{\times}8$ pixel neighborhood and use it as a geometric signal for stroke ordering and scale: regions with lower local variation are assigned earlier and larger strokes, while regions with greater variation receive later and smaller strokes.

\textit{Attention Map.}
DINO self-attention provides a complementary perceptual signal~\cite{caron2021emergingpropertiesselfsupervisedvision,amir2022deepvitfeaturesdense}. We use last-layer center-patch attention from a ViT-S/16 model as a spatial sampling weight, allocating more stroke samples to locations with higher attention values.

Regions are processed in coarse-to-fine label order. Within each region, $N$ stroke positions are sampled from the DINO attention map and scored by $\text{score}_i=100\cdot\hat{\sigma}_i+\hat{a}_i$, where $\hat{\sigma}_i$ and $\hat{a}_i$ denote normalized local normal variation and attention, respectively. Sorting this score places flatter, less-salient positions earlier. Brush sizes follow a linear decay from a region-specific maximum to minimum. Once assembled, stroke parameters are jointly optimized against the target image through differentiable rendering using Adam with cosine annealing over 30 iterations and early stopping.

\subsubsection{Target Predictor}
\label{sec:target_predictor}

The Target Predictor estimates a plausible intended outcome from the current canvas state alone. All intermediate canvas states from the same painting sequence share the completed painting as their supervision target, forming a many-to-one prediction task. The model can therefore update its estimate as additional visual information accumulates on the canvas without requiring the artist to provide an explicit target image.

We implement the Target Predictor as a compact Vision Transformer that directly regresses a target latent in a single forward pass rather than sampling a completion through an iterative generative process. Because the prediction is repeatedly updated as the canvas evolves, a single-pass model provides a stable conditioning signal without requiring multi-step sampling. A deterministic estimate is also useful downstream because PaintCopilot's optimization-based workflows operate against a fixed target during each intervention.

The model operates in the latent space of a pretrained Stable Diffusion VAE~\cite{rombach2022high}. The canvas latent is patchified into tokens with learned positional embeddings and processed by a lightweight Vision Transformer. The model outputs the predicted target latent $\hat{z}$ directly from the canvas latent $z_c$ and is supervised with a combined L1 and MSE objective, $\mathcal{L}_{\mathrm{target}}=0.8\,\|\hat{z}-z^*\|_1+0.2\,\|\hat{z}-z^*\|_2^2$, where $z^*$ is the VAE-encoded ground-truth target. Small Gaussian noise is added to $z^*$ during training as regularization against overfitting to exact latent values.

The inferred target is not treated as a commitment to a fixed direction. In the interface, artists can inspect the current estimate and replace it with an explicit reference when desired. The prediction therefore provides a revisable representation of direction rather than a predefined endpoint imposed before painting begins.

\subsubsection{Stroke Predictor}
\label{sec:next_stroke}

The Stroke Predictor generates a plausible next brushstroke conditioned on the current canvas state and a window of prior strokes. It outputs an 8-dimensional parameter vector $a=(x,y,p,r,\theta,R,G,B)$, encoding position, pressure, size, angle, and color. We model strokes directly as continuous vectors through flow matching rather than through a discrete codebook or token representation~\cite{ha2017neural}. The continuous representation preserves generated outputs as explicit painting actions that can enter the same rendering and optimization pipeline as existing strokes.

The canvas VAE latent is first compressed into a fixed-dimensional condition vector $c$ by a small MLP encoder. Prior strokes are projected into token embeddings with learned positional encodings and processed by a causal Transformer conditioned on $c$ through Adaptive Layer Normalization (AdaLN). The model conditions on the 50 most recent strokes, keeping the self-attention context bounded as the painting progresses. Causal masking restricts each position to preceding strokes so that training matches the information available during inference. The final token representation $h$ provides context for stroke generation. Conditioning jointly on the current canvas and recent actions supports history-dependent continuation at the level of individual strokes.

Because multiple strokes may plausibly follow the same context, we formulate generation as a Flow Matching problem~\cite{lipman2023flow} rather than regressing toward a single next-stroke estimate. We learn a velocity field $v_\theta(a_t,h,t)$ that transports samples from a standard Gaussian prior $a_0\sim\mathcal{N}(0,I)$ toward the target stroke distribution. The training path and target velocity are $u_t=a_{\mathrm{tar}}-a_{\mathrm{src}}$ and $a_t=(1-t)a_{\mathrm{src}}+ta_{\mathrm{tar}}$, and the model minimizes $\mathcal{L}_{\mathrm{FM}}=\mathbb{E}_{t,a_{\mathrm{src}},a_{\mathrm{tar}}}\left[\left\|v_\theta(a_t,h,t)-u_t\right\|_2^2\right]$.

At inference time, a stroke is sampled by integrating the learned velocity field from $t=0$ to $t=1$ using Euler steps. Modeling a distribution over subsequent strokes reflects the open-ended nature of continuation, in which the same painting state may admit multiple plausible next actions. Repeated sampling allows the model to extend the painting autoregressively while updating its context after each generated stroke.

\subsubsection{Region Sampler}
\label{sec:region_sampler}

The Region Sampler generates a stroke sequence within an artist-specified region under a semantic condition. Rather than determining every stroke individually, the artist specifies \emph{where} generation may occur through a bounding region and \emph{what} type of content should be generated through a semantic label. This provides a complementary level of continuation in which delegation is spatially and semantically bounded.

The model is a Variational Autoencoder operating on sequences of stroke parameters. The encoder prepends a learnable CLS token to the stroke sequence, processes the sequence with a Transformer, and projects the CLS representation to a Gaussian posterior through separate linear heads. The decoder conditions on both the sampled latent $z$ and a semantic-label embedding, broadcasts the combined representation to a fixed-length token sequence, and reconstructs stroke parameters through a Transformer followed by a sigmoid output projection.

Stroke positions are represented in local bounding-box coordinates normalized to the region extent, making the representation independent of the region's absolute position and scale on the canvas. The same learned stroke distribution can therefore be sampled within artist-selected regions of different sizes and locations.

The model is trained with the ELBO objective
\begin{equation*}
\small
\mathcal{L}_{\mathrm{VAE}} =
\mathbb{E}_{q(z|x)}[\|x-\hat{x}\|_2^2]
+\beta D_{\mathrm{KL}}(q(z|x)\,\|\,\mathcal{N}(0,I)),
\end{equation*}
where $\beta$ is annealed from zero to a small value during training.
Sampling from the resulting latent space allows different stroke arrangements to be generated under the same semantic and spatial constraints rather than requiring a single deterministic sequence.

\subsubsection{Training Details}
\label{sec:training}

All models are trained on 8 NVIDIA RTX Pro 6000 GPUs using AdamW with gradient clipping at 1.0 and cosine-annealing learning-rate schedules with linear warmup. The Target Predictor and Stroke Predictor are each trained for 500 epochs with a learning rate of $10^{-4}$ and batch sizes of 256 and 512, respectively. The Region Sampler is trained for 5{,}000 epochs with a learning rate of $10^{-3}$ and batch size 128, with $\beta$ annealed linearly from $10^{-4}$ to $10^{-3}$ over the first 2{,}000 epochs.

\begin{figure}
    \centering
    \includegraphics[width=\linewidth]{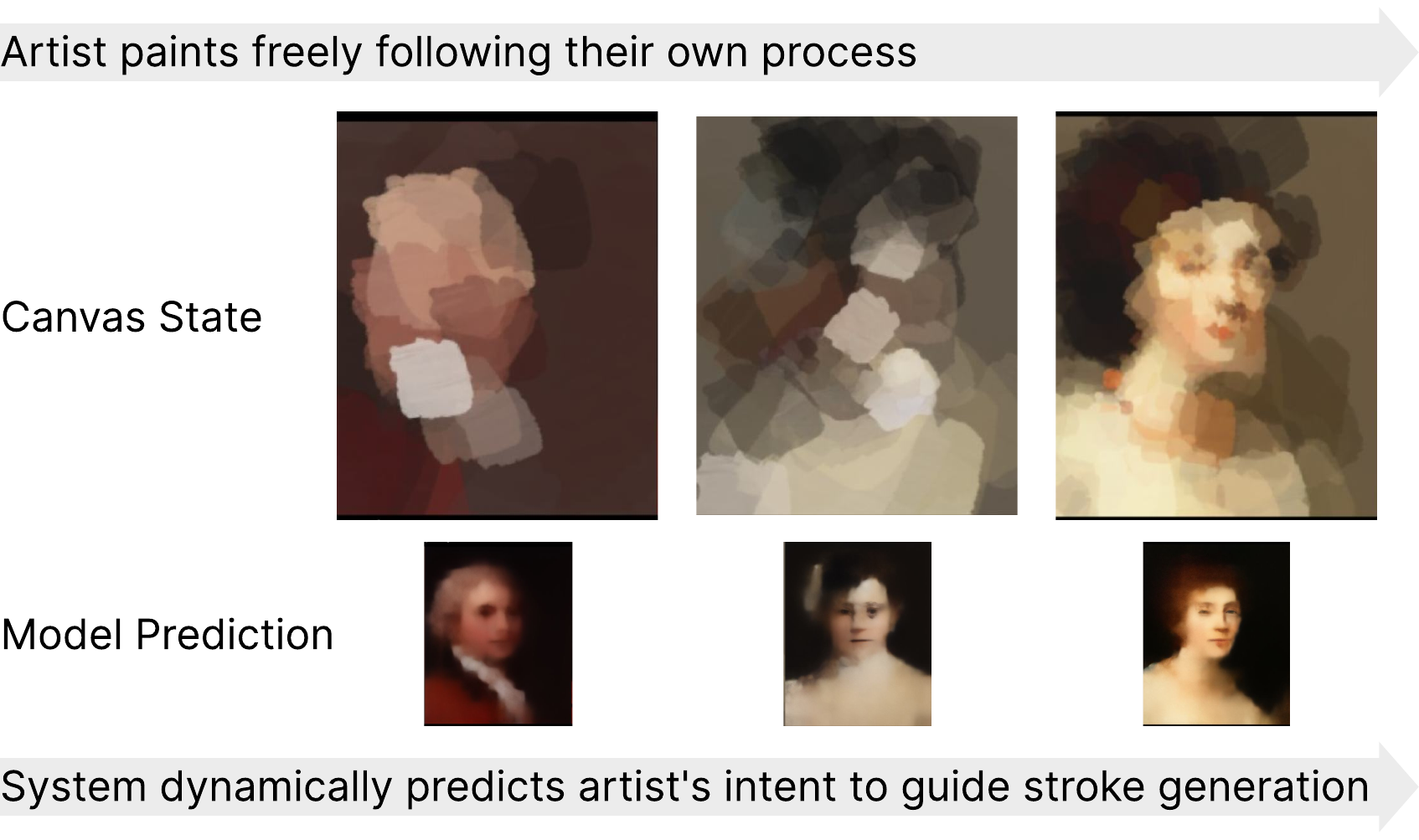}
    \caption{Evolving target prediction during painting. As the canvas develops, PaintCopilot updates its estimate of artistic direction, providing a revisable conditioning signal for subsequent AI interventions rather than requiring a predefined target.}
    \label{fig:target_evolution}
\end{figure}

\subsection{Implementation}
\label{sec:implementation}

PaintCopilot is implemented as a web application with a browser-based frontend written in HTML, CSS, and JavaScript and a FastAPI backend for model inference and differentiable stroke optimization. The frontend and backend communicate over HTTP, with streaming NDJSON responses used to return intermediate results during Stroke Completion, Region Inpainting, and Optimize History. Model inference and differentiable optimization are implemented in PyTorch with support for CUDA, MPS, and CPU execution. In our deployment, inference was hosted on a remote GPU server and accessed through the browser-based interface, allowing participants to use PaintCopilot without installing or running the models locally.

\section{Technical Evaluation}
We evaluate PaintCopilot from two complementary perspectives. We first examine the basic behavioral properties of each learned component. 

\subsection{Component-Level Evaluation}
We evaluate all three components on a common held-out set of 100 paintings sampled from outside the training set. Because autonomous continuation admits multiple plausible next actions from the same painting state, and the ground-truth targets and stroke sequences are themselves produced by our stroke-based rendering pipeline rather than recorded from human painters, these evaluations do not measure artistic validity. Instead, they provide basic behavioral checks: whether predictions improve with additional visual evidence, sampled strokes remain close to plausible continuations rather than diverging arbitrarily, and generation avoids collapsing to a single degenerate output. Whether generated continuations are perceived as artistically valid by practicing artists is addressed separately through the deployment study in \autoref{sec:user_study}.

\begin{figure}
    \centering
    \includegraphics[width=\linewidth]{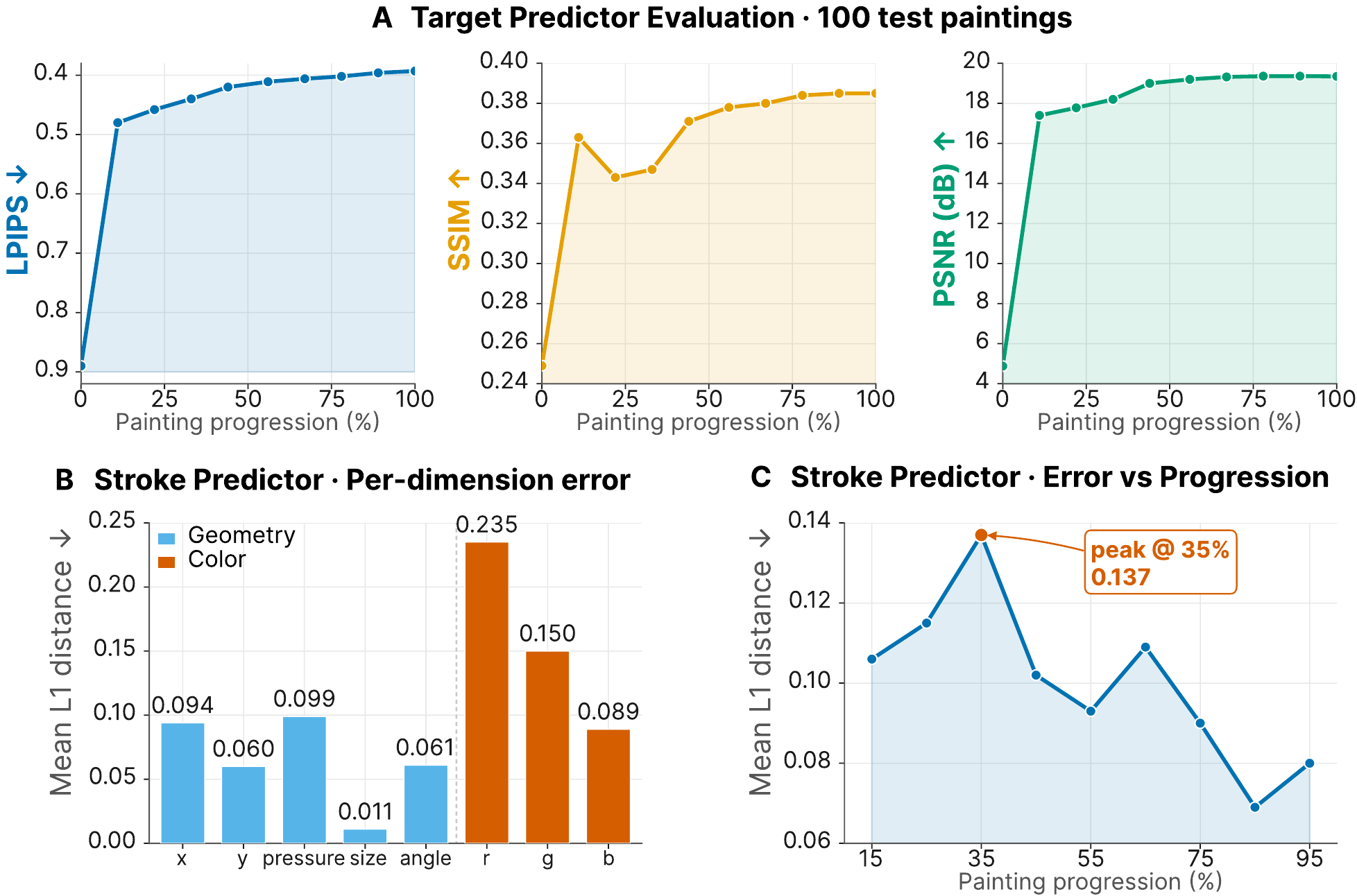}
    \caption{Quantitative evaluation of the Target Predictor and Autoregressive Stroke Predictor. (a): Target prediction quality across painting progression, measured by LPIPS, SSIM, and PSNR; (b): Mean L1 prediction error for each stroke parameter; (c): Mean L1 prediction error across painting progression.}
    \label{fig:eval}
\end{figure}

\noindent\textbf{Target Predictor.}
For each held-out painting, we measure prediction quality against the ground-truth target image at ten evenly spaced stages of painting progression. We report LPIPS~\cite{zhang2018unreasonableeffectivenessdeepfeatures} (lower is better), SSIM (higher is better), and PSNR (higher is better) between the predicted and ground-truth target images decoded from VAE latent space. Prediction accuracy improves consistently with painting progression (\autoref{fig:eval}(a)). At the blank canvas stage, LPIPS is 0.89 and PSNR is 4.88 dB, reflecting the difficulty of inferring a specific target without visual evidence. After just 11\% of strokes are laid down, LPIPS drops sharply to 0.48 and PSNR rises to 17.40 dB, indicating that early compositional strokes already provide substantial information for target prediction.

\noindent\textbf{Stroke Predictor.}
Given a window of 50 prior strokes and the current canvas state, the Stroke Predictor samples the next stroke as an 8-dimensional parameter vector. We use mean L1 distance between sampled and ground-truth stroke parameters as a reference measure of local continuation fidelity, reported overall, per dimension, and across painting progression. The model achieves an overall mean L1 distance of 0.100 across all dimensions. Per-dimension analysis (\autoref{fig:eval}(b)) shows the highest prediction error for color channels and the lowest for brush size. Prediction error also varies with painting progression (\autoref{fig:eval}(c)), peaking mid-process and decreasing through the latter half of the painting. Because multiple next strokes may be valid for the same context, these errors characterize proximity to the recorded continuation rather than a uniquely correct or artistically valid prediction.

\noindent\textbf{Region Sampler.}
We evaluate the Region Sampler along two axes. For reconstruction quality, we encode and decode each stroke chunk per semantic label through the VAE and measure mean L1 distance between the reconstructed and original stroke sequences. For generation diversity, we sample 10 random latent codes per semantic label and compute mean pairwise L1 distance across the decoded stroke sequences. The Region Sampler achieves an overall reconstruction mean L1 of 0.179. Reconstruction error varies across semantic labels, with fine-grained facial regions such as hair and eye detail showing lower error than coarser regions such as background and hat. Across all evaluated labels, samples exhibit non-zero pairwise variation, indicating that the model generates different stroke arrangements rather than reproducing a single sequence.

Together, these results establish the basic behavioral properties required by our formulation: the Target Predictor improves as painting context accumulates, the Stroke Predictor samples continuations that remain close to held-out trajectory behavior rather than drifting arbitrarily, and the Region Sampler reconstructs held-out stroke sequences while producing non-identical samples. These are necessary conditions for the interactive workflows built on top of them; whether they are useful to artists in practice is examined next through the deployment study.

\begin{figure*}
    \centering
    \includegraphics[width=1\linewidth]{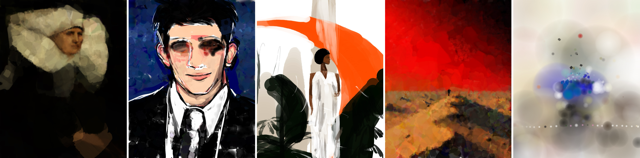}
    \caption{Artworks created during the two-week deployment illustrate different strategies for incorporating PaintCopilot into individual painting practices. Some participants relied extensively on AI continuation, while others primarily painted manually and invoked AI only for selected parts of the process. Participants also explored the system beyond its portrait-focused training domain, using it with different subjects and visual styles.}
    \label{fig:placeholder}
\end{figure*}

\section{User Study}
\label{sec:user_study}

We conducted a two-week deployment study to understand how artists used PaintCopilot in their own creative practice and how its autonomous assistance affected their sense of agency and ownership. Rather than evaluating individual features through prescribed drawing tasks, we allowed participants to decide what to paint and when, where, and how much to involve PaintCopilot. Our study consists of two parts. \textbf{Part 1 (Using Autonomous Assistance)} examines how participants used and controlled PaintCopilot's autonomous assistance during painting, including how they set the target, used different intervention workflows across the painting process, and decided where and how long the system could act. \textbf{Part 2 (Agency and Ownership)} examines participants' perceived control and ownership when creating with PaintCopilot, together with how they perceived the system's contribution and alignment with their creative direction.

\subsection{Study Design}

\noindent\textbf{Participants.}
We recruited ten participants between 23 and 28 years old (\(M = 25.30\), \(SD = 1.57\), \(Mdn = 25.5\)). Five identified as women, four as men, and one preferred not to disclose their gender. Participants had between 0 and 26 years of drawing experience (\(M = 12.50\), \(Mdn = 11\)). Three participants identified as professional artists, and five had previously used generative AI for drawing or painting. We intentionally recruited participants with varying levels of drawing experience, professional involvement in art, and familiarity with generative AI to examine PaintCopilot across different existing creative practices. Participants were required to have access to either a pressure-sensitive drawing tablet or an iPad with Apple Pencil, as well as sufficient time for personal creative work during the deployment. The study was approved by our institution's ethics review board, and all participants provided informed consent.

\noindent\textbf{Procedure.}
The study consisted of a 60-minute entrance session, a two-week deployment, and a 30-minute exit interview. During the entrance session, we introduced PaintCopilot and guided participants through its four intervention workflows---Dynamic Brush, Optimize History, Region Inpainting, and Stroke Completion---as well as its predicted and explicitly specified target modes. Participants then used PaintCopilot in their normal creative environments for two weeks and completed at least three paintings of their own choosing. We did not prescribe what they should paint or how they should use AI assistance, allowing different patterns of use to emerge naturally. Throughout the deployment, we collected interaction logs, screen recordings, reflection forms, and resulting artworks.

The exit interview connected participants' actual use of PaintCopilot to their reflections on the experience. Before each interview, we reviewed their interaction records and artworks and identified specific episodes for follow-up. In \textbf{Part 1 (Using Autonomous Assistance)}, participants walked us through how they involved PaintCopilot in their paintings: how they worked with targets, chose among intervention workflows at different stages, constrained where the system could act, and decided when to take over. Building on these concrete interactions, \textbf{Part 2 (Agency and Ownership)} shifted from how participants used AI assistance to how they experienced it, examining their perceived control and ownership, PaintCopilot's contribution, and its alignment with their intended creative direction.

\noindent\textbf{Data Collection and Analysis.}
We collected participants' interaction logs, screen recordings, reflection forms, artworks, questionnaire responses, and exit-interview responses. We used interaction records and artworks to contextualize participants' retrospective accounts and identify specific episodes for follow-up. For Part 1, we summarized target-setting strategies, workflow use across painting stages, spatial constraints, and Stroke Completion takeover behavior using descriptive statistics, and examined participants' accounts alongside these behaviors to characterize recurring ways in which they adjusted AI participation. For Part 2, we summarized the four seven-point ratings and reviewed participants' accounts of authorship, control, ownership, and future use alongside their recorded interactions and artworks.

\subsection{Results}

\begin{figure}[t]
    \centering
    \includegraphics[width=1\columnwidth]{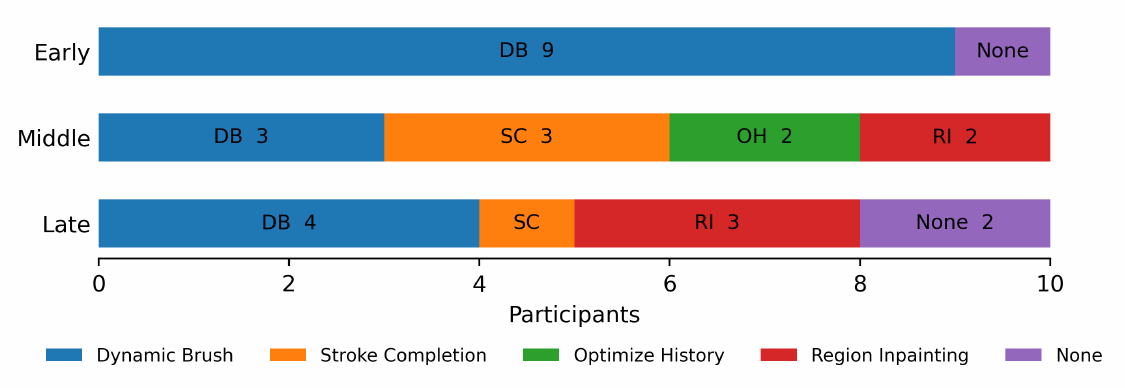}
    \caption{
    Participants' most frequently used PaintCopilot workflow during the early, middle, and late stages of painting. Dynamic Brush dominated early-stage use, while workflow use became more varied during the middle and late stages. DB = Dynamic Brush; SC = Stroke Completion; OH = Optimize History; RI = Region Inpainting. ``None'' indicates that the participant reported using none of the four workflows during that stage.
    }
    \label{fig:workflow_stage}
\end{figure}

\noindent\textbf{Using Autonomous Assistance.}
Participants did not adopt a single fixed level of AI involvement. Instead, their use of PaintCopilot varied along three recurring dimensions: the \emph{direction} guiding generation, the \emph{scope} of AI intervention, and the \emph{persistence} of generated contributions. Five of ten participants primarily set targets manually, two primarily relied on system prediction, one used both approaches about equally, and two did not use target prediction. When predictions differed from their intentions, participants often continued painting, modified the canvas to steer subsequent predictions, or manually specified a target. For example, when PaintCopilot interpreted P3's landscape as a portrait, P3 continued adding landscape details until the prediction gradually shifted toward the intended scene; P6 similarly added darker tones and more explicit contours to steer an undesired prediction toward their intended subject. Participants also changed workflows as their paintings developed (\autoref{fig:workflow_stage}). Dynamic Brush dominated the early stage (9/10), while use became more varied in the middle stage, with Dynamic Brush and Stroke Completion each reported by three participants and Optimize History and Region Inpainting by two each. In the late stage, Dynamic Brush (4/10) and Region Inpainting (3/10) were most common, while two participants used none of the four workflows.

Participants also differed in how they bounded autonomous assistance. For spatial constraints, the nine participants who rated whether lasso selection made them more willing to allow larger changes reported a mean rating of 3.78 (\(SD=1.20\)) on a seven-point scale. Some valued the added precision, whereas others felt that generation remained unpredictable even within a selected region. For Stroke Completion, three participants did not use the feature, four reported no consistent number of strokes before takeover, and the remaining three reported 4--5, 6--10, and more than 10 strokes, respectively. Takeover was often prompted by the generated content rather than a fixed number of strokes. For example, P6 stopped the system after it covered previously drawn facial details, while P7 took over when generated strokes turned a sketch into large, blurred regions of color. Overall, participants adjusted autonomous assistance through target setting, workflow selection, spatial constraints, and takeover rather than following a single fixed pattern of use.

\begin{figure}[t]
    \centering
    \includegraphics[width=1\columnwidth]{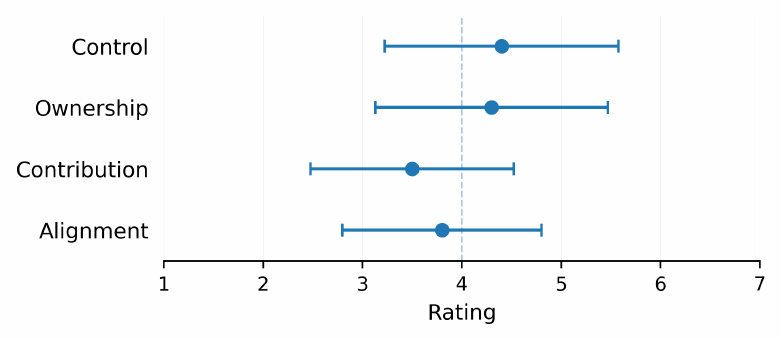}
    \caption{
    Participants' ratings of perceived control, ownership, PaintCopilot's contribution, and alignment with their intended direction after the two-week deployment. Dots indicate means and error bars indicate 95\% confidence intervals on seven-point scales; the dashed line marks the scale midpoint.
    }
    \label{fig:agency_ownership}
\end{figure}

\noindent\textbf{Agency and Ownership.}
Participants reported mixed experiences of agency and ownership when creating with PaintCopilot. As shown in \autoref{fig:agency_ownership}, ratings varied considerably across participants: Control averaged 4.40 (\(SD=1.90\)), Ownership 4.30 (\(SD=1.89\)), Contribution 3.50 (\(SD=1.65\)), and Alignment 3.80 (\(SD=1.62\)) on seven-point scales. Participants' accounts reflected similar differences in perceived authorship. P3 regarded the resulting paintings as their own because PaintCopilot mainly added details and textures rather than determining the main content, while P2 emphasized that the main ideas, decisions, and corrections remained theirs. In contrast, P7 regarded the results as AI-influenced rather than completely their own, and P6 described them as collaborative works that reflected their intentions but were not fully created by themselves.

Participants likewise differed in whether and how they actively maintained ownership during painting. Some frequently revised PaintCopilot's contributions: P5 redrew generated regions and preserved their own strokes, P3 manually completed important elements, and P2 undid or painted over outputs that did not match their intentions. Others did not deliberately try to preserve a distinct sense of authorship; P6, for example, emphasized the collaborative process rather than claiming the resulting work as solely their own. Future adoption was similarly mixed: five participants said they would probably or definitely use PaintCopilot in their regular creative practice, two were unsure, and three said they probably or definitely would not. Together, these responses show that autonomous assistance led to different relationships with authorship, ranging from using PaintCopilot as a bounded tool within largely self-directed work to viewing the resulting painting as a more collaborative or AI-influenced outcome.

\subsection{User Feedback}

Participants distinguished PaintCopilot from prompt-based image-generation tools primarily by how AI entered the creative process. Rather than repeatedly prompting, waiting for a complete result, and revising the prompt, participants could respond to AI assistance while continuing to paint. P5 described this distinction as a shift from iterative prompting to incremental co-creation:

\begin{quote}
\emph{``Other AI tools directly give me a complete result, which I then have to revise step by step. With PaintCopilot, I can work with the AI step by step instead of getting everything at once.''}
\end{quote}

P6 similarly contrasted PaintCopilot with Stable Diffusion and GPT-based image generation, describing those tools as a back-and-forth process that required waiting for each result, whereas PaintCopilot felt more like modifying the painting in real time:

\begin{quote}
\emph{``Stable Diffusion and GPT are more back-and-forth---you have to wait for the generated result. PaintCopilot feels more like making changes in real time, so it is easier to adjust.''}
\end{quote}

P8 emphasized a related difference in how ideas emerged during the creative process. Rather than instructing an image generator and responding to its completed output, PaintCopilot could introduce suggestions while the artist was actively realizing their own ideas:

\begin{quote}
\emph{``When I draw by hand, I start thinking from the perspective of painting. While realizing my own ideas, AI can provide new inspiration and directions.''}
\end{quote}

This process-oriented interaction was particularly valued during exploratory stages. P6 described using PaintCopilot during moodboard exploration, when the desired details were not yet precisely defined, while P5 preferred it when having ``a basic idea that is not yet clear'' and wanting AI assistance to develop an initial sketch. P2 similarly preferred early exploration, when the direction was still open and AI suggestions could become part of the trial-and-error process. Together, these accounts suggest that participants valued PaintCopilot not simply for generating visual content, but for allowing suggestions to emerge and be adjusted within the ongoing act of painting.

However, participants also emphasized that this form of intervention became less useful when their intentions were already precise. P2 found PaintCopilot's continuation-based interaction well suited to impasto-like painting, where colors and forms are continuously layered and adjusted, but less compatible with anime-style drawing, where contours, color boundaries, and predefined shapes require tighter control:

\begin{quote}
\emph{``The biggest problem is that it does not always know how much I want it to help at that moment. For styles like anime that require clear shapes and control, it is easier to feel that the AI is not listening.''}
\end{quote}

Other participants reported similar limitations when generated strokes overwrote existing details, introduced unexpected colors, or failed to follow the intended subject. P3 appreciated that PaintCopilot could intervene through brushstrokes with adjustable levels of intervention, but found Region Fill less effective than comparable image-generation tools. P4 also noted that generation latency could interrupt the natural rhythm of drawing. Overall, participants valued PaintCopilot's ability to bring AI assistance into the ongoing act of painting, particularly when they were exploring possibilities and willing to accommodate unexpected suggestions, while more precisely specified intentions demanded tighter control over AI intervention.

\section{Discussion}

PaintCopilot began from a technical question: whether painting could be modeled as continuation rather than reconstruction. Our deployment suggests that this reformulation also changes the interaction problem. Once AI can propose and execute plausible next actions without a fixed target, control is no longer only about specifying a desired output. Participants instead exercised control throughout the process by redirecting inferred targets, choosing forms of assistance, constraining where generation occurred, interrupting continuation, and revising generated strokes.

\noindent\textbf{Negotiating AI Participation.}
The appropriate degree of AI intervention changed as participants' intentions developed. Unexpected suggestions were often welcomed during early exploration, when the desired outcome remained open, but could become disruptive once contours, details, or stylistic decisions became more precise. Across these interactions, AI participation was continually adjusted rather than fixed. Participants changed the direction guiding AI contributions, the spatial and temporal scope of its intervention, and the persistence of its outputs through correction, revision, or overwriting. This allowed AI to act without requiring approval of every generated action while leaving the boundaries of its participation open to revision. The design question therefore shifts from determining how much autonomy AI should have to supporting creators in continually adjusting AI participation as their intentions evolve.

\noindent\textbf{Autonomous Generation as Revisable Material.}
Participants frequently painted over generated strokes, preserved useful portions, or developed them further rather than simply accepting or rejecting a result. Because PaintCopilot generates explicit brushstrokes in the artist's action space, its output remains provisional within the painting process. We therefore distinguish between \emph{generated results} and \emph{generated material}: the former invites evaluation of whether an output is satisfactory, whereas the latter gains value through how it can be transformed and incorporated into subsequent actions. For process-oriented creative systems, revisability may therefore be as important as generating an immediately correct result.

\noindent\textbf{Creative Agency Through Selective Delegation.}
Participants' mixed experiences of control and ownership suggest that creative agency cannot be explained simply by how much content AI produces. Some regarded the work as their own when PaintCopilot elaborated within a direction they established, while others reported weaker ownership when it determined colors, forms, or other decisions central to the work. Participants reclaimed such decisions by undoing, painting over, constraining, or taking over from the system. We therefore view agency as \emph{selective delegation}: the capacity to decide which creative decisions can be delegated and reclaim them as they become consequential. Process-oriented co-creative systems should thus support autonomy not as a fixed division of labor, but as a relationship that creators can continually renegotiate as their work and intentions evolve.

\section{Limitations and Future Work}

\noindent\textbf{Study Scope.}
Our deployment involved ten participants using PaintCopilot for two weeks. The self-directed design allowed participants to incorporate the system into their own creative practice and develop different strategies for delegation and revision, but does not support controlled comparisons with alternative forms of AI assistance. Longer and comparative studies could examine how these strategies develop with familiarity and whether process-level continuation produces different experiences from image-level generation or editing.

\noindent\textbf{Generalizability.}
Our training data and technical evaluation are restricted to portrait painting, with stroke sequences generated through a rendering pipeline rather than recorded from real painters. Our implementation also represents digital painting through three differentiable brush representations. Whether the architecture and training pipeline transfer to more open-ended compositions, different painting styles, or human-recorded sequences remains untested. Future work could extend autonomous continuation to richer action spaces involving layers, erasing, compositing, different brushes, or physical-media processes.

\noindent\textbf{Representing Artistic Intention.}
PaintCopilot models continuation from the current canvas and a limited history of preceding strokes, while artistic intention may also depend on conceptual goals, external references, stylistic precedents, and plans for later stages. The Target Predictor represents evolving direction, but participants' experiences showed that its inference did not always match their intention at a particular moment. Future systems could investigate richer, explicitly editable representations through which artists communicate, inspect, and revise the context guiding autonomous continuation.

\noindent\textbf{Evaluating Artistic Continuation.}
Our technical evaluation verifies whether the learned components exhibit the behaviors needed for PaintCopilot's workflows rather than establishing a unique notion of correct artistic continuation. Multiple actions may be plausible from the same painting state, and similarity to a reference continuation cannot by itself determine artistic validity. Future evaluation should account for diversity and contextual appropriateness, as well as whether generated actions are steerable, revisable, and useful for subsequent human decisions.

\section{Conclusion}
We presented PaintCopilot, a co-creative neural painting system built around \emph{autonomous artistic continuation}: generating plausible subsequent painting actions from an evolving canvas and stroke history rather than reconstructing a predefined target. A two-week deployment with ten artists showed that this capability did not produce a simple handoff between human and AI. Participants' accounts suggest that their sense of agency was shaped less by the amount of work delegated than by their ability to continually adjust AI participation. AI could act autonomously within the creative process while artists retained the ability to redirect its interpretation, constrain where and when it intervened, and revise or overwrite its contributions as their intentions evolved.


\balance

\bibliographystyle{ACM-Reference-Format}
\bibliography{references}

\end{document}